\pdfoutput=1
\documentclass[10pt,twocolumn,letterpaper]{article}

\usepackage{cvpr}
\usepackage{times}
\usepackage{epsfig, mathtools}
\usepackage{graphicx}
\usepackage{amsmath}
\usepackage{amssymb}

\usepackage{booktabs}
\usepackage{tabulary}
\usepackage[dvipsnames]{xcolor}
\usepackage[caption=false]{subfig}
\usepackage[british,english,american]{babel}
\usepackage{soul}
\usepackage{xspace}\xspace
\usepackage[export]{adjustbox}
\usepackage{array}
\usepackage{enumitem}
\usepackage{dblfloatfix}
\usepackage{transparent}
\usepackage{subfig, float, multirow}
\usepackage[]{algorithm2e}
\usepackage{listings}
\usepackage[symbol]{footmisc}
\usepackage{etoolbox}

\newcommand{\invenio}[0]{\texttt{Invenio} }
\patchcmd{\maketitle}
{\def\@makefnmark}
{\def\@makefnmark{}\def\useless@macro}
{}{}

\definecolor{citecolor}{HTML}{0071bc}
\usepackage[pagebackref=false,breaklinks=true,letterpaper=true,colorlinks,citecolor=citecolor,bookmarks=false]{hyperref}

\newcommand{\app}{\raise.17ex\hbox{$\scriptstyle\sim$}}

\newcolumntype{x}[1]{>{\centering\arraybackslash}p{#1pt}}
\newcolumntype{y}[1]{>{\raggedright\arraybackslash}p{#1pt}}
\newcolumntype{z}[1]{>{\raggedleft\arraybackslash}p{#1pt}}
\newlength\savewidth

\makeatletter\renewcommand\paragraph{\@startsection{paragraph}{4}{\z@}
  {.5em \@plus1ex \@minus.2ex}{-.5em}{\normalfont\normalsize\bfseries}}\makeatother

\cvprfinalcopy 

\begin{document}

\title{Invenio: Discovering Hidden Relationships Between Tasks/Domains Using Structured Meta Learning \vspace{-.5em}\thanks{The first two authors contributed equally.}}
	
\author{
\vspace{.5em}
 Sameeksha Katoch$^{\dagger *}$ \quad  Kowshik Thopalli$^{\dagger *}$ \quad Jayaraman J. Thiagarajan$^{\ddagger}$ \quad\\ Pavan Turaga$^{\dagger}$ \quad Andreas Spanias$^{\dagger}$ \vspace{.5em}\\
 $^{\dagger}$Arizona State University, $^{\ddagger}$Lawrence Livermore National Labs \vspace{.3em}}
\maketitle

\begin{abstract}
\vspace{-.5em}
Exploiting known semantic relationships between fine-grained tasks is critical to the success of recent model agnostic approaches. These approaches often rely on meta-optimization to make a model robust to systematic task or domain shifts. However, in practice, the performance of these methods can suffer, when there are no coherent semantic relationships between the tasks (or domains). We present\textnormal{~\texttt{Invenio}}, a structured meta-learning algorithm to infer semantic similarities between a given set of tasks and to provide insights into the complexity of transferring knowledge between different tasks. In contrast to existing techniques such as Task2Vec and Taskonomy, which measure similarities between pre-trained models, our approach employs a novel self-supervised learning strategy to discover these relationships in the training loop and at the same time utilizes them to update task-specific models in the meta-update step. Using challenging task and domain databases, under few-shot learning settings, we show that \textnormal{\texttt{Invenio}} can discover intricate dependencies between tasks or domains, and can provide significant gains over existing approaches in terms of generalization performance. The learned semantic structure between tasks/domains from \textnormal{\texttt{Invenio}} is interpretable and can be used to construct meaningful priors for tasks or domains. 

\vspace{-1em}
\end{abstract}
\section{Introduction}
\label{sec:intro}
The success of deep learning in a wide-variety of AI applications can be partly attributed to its ability to be re-purposed for novel tasks or operating environments. This is particularly crucial in data-hungry scenarios, e.g. few shot learning, where datasets or models from related tasks can be effectively leveraged to solve the task at hand. Though \textit{transfer learning} methods have been proposed for applications including computer vision~\cite{Zamir2018TaskonomyDT}, language processing~\cite{Devlin2019BERTPO,Yang2019XLNetGA} and medical image analysis~\cite{khan2019machine}, even the more sophisticated approaches (based on deep neural networks) are often found to be brittle when applied in scenarios characterized by challenging domain and task shifts. Consequently, it is imperative to qualify the degree of similarity between the training and testing scenarios (domain or task), in order to assess if a model can be effectively re-purposed. This naturally calls for approaches that can reason about the semantic space of tasks (or domains), and to quantify how difficult it is to transfer from a scenario to another. In this spirit, Achille \textit{et al.}~\cite{achille2019information} recently proposed an information-theoretic framework for characterizing the complexity of tasks through the information from parameters of a deep neural network, and designed an asymmetric distance function between tasks that was showed to be strongly correlated to the ease of transfer learning based on model fine-tuning. A similar distance function was utilized by~\texttt{TASK2VEC}~\cite{Task2Vec} to produce embeddings that describe the semantic space of tasks. More specifically, this distance is computed based on Fisher information from parameters of trained networks for different tasks, and hence this implicitly assumes access to sufficient training data. As a result, it is not suitable for comparing tasks in few-shot learning scenarios.

In this paper, we represent both tasks and domains by few-shot datasets of input images and discrete output labels, and our goal is to infer hidden relationships between tasks or domains. We present~\texttt{Invenio}, a scalable model agnostic (meta learning) approach, that can effectively infer the semantic structure of the space of tasks (or domains) and at the same time leverage the inferred relationships to perform task-specific model optimization. In other words, instead of explicitly performing model fine-tuning between two similar tasks, \texttt{Invenio} identifies the structure that is maximally beneficial for transfer learning between the entire set of tasks or domains. 

\begin{figure*}
	\centering
	\includegraphics[width=0.8\textwidth]{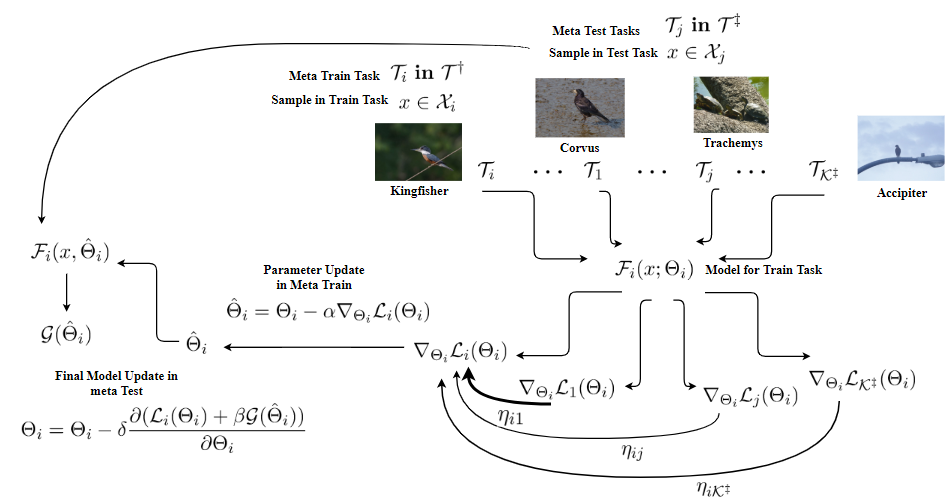}
	\caption{An overview of the proposed structured meta learning technique for the case of tasks. Each task is assumed to have separate model parameters $\Theta_i$ and this illustration shows the parameter update for one train task. This formulation directly applies to the case of domains as well.}
	\label{fig:block_diag}
	\vspace{-10pt}
\end{figure*}
\vspace{10pt}
\noindent \textbf{Proposed Work.} The recent class of meta-optimization approaches for few-shot domain/task generalization~\cite{Finn2017ModelAgnosticMF, Li2017LearningTG} attempt to learn a single model on a set of observed tasks, which is assumed to be only a few gradient descent steps away from good task-specific models. Their success hinges on the assumption that the observed set of tasks or domains are realizations from a common distribution. However, in practice, the degree of
similarity between tasks or domains are unknown \textit{a priori}, and hence the assumption of finding a single base learner could be restrictive. In contrast,~\texttt{Invenio} makes a general assumption that there exists an inherent semantic space of tasks/domains, wherein information from each subset of related tasks (or domains) can be used to make a task-specific learner effective. To this end, we develop a structured meta-learning algorithm (Figure~\ref{fig:block_diag}) that infers semantic similarities between different tasks (or domains), while also obtaining generalizable base learners. More specifically, our approach allows each task (or equivalently domain) to use separate model parameters while enabling information sharing between related tasks, and trains them for generalization using gradient-through-gradient style optimization. A crucial outcome from the proposed approach is a structured semantic space of tasks (or domains) which can be utilized to build powerful task/domain priors. In order to demonstrate the use of \texttt{Invenio}, we design challenging task ($400$ tasks) and domain ($53$ domain shifts) databases and show that the inferred semantic relationships are highly interpretable. More importantly, \texttt{Invenio} provides significant performance gains, in terms of generalization, when compared to conventional strategies.



\vspace{0.1in}

\noindent Our contributions can be summarized as follows:
\begin{itemize}[noitemsep,topsep=0pt,parsep=0pt,partopsep=0pt]
	\item Unfolding the inherent semantic structure for tasks/domains in few-shot learning settings, which can be used for designing effective priors.
	\item A structured meta learning algorithm for leveraging the similarities to build highly generalizable models.
	\item Empirical studies with custom task and domain databases to show the effectiveness of \texttt{Invenio} to identify meaningful relationships which can be used to enable improved generalization. 
\end{itemize}

\section{Problem Setup}
\label{sec:problem}
In this paper, we consider systematic task and domain shifts in image classification, and explore the use of a model agnostic approach for inferring semantic similarities between them using few-shot datasets. We begin by describing the overall setup and assumptions.

We represent each task or domain as a few-shot dataset with input images and output space of labels. We denote the set of $K$ labeled datasets corresponding to a set of tasks by the general notation $\mathcal{T}_i \coloneqq \{\mathrm{X}_i, \mathrm{Y}_i\}, i=1,\cdots,K$. Each domain $\mathcal{D}_i$ is also defined similarly using a finite dataset. We consider a few-shot setting, wherein each dataset is comprised of $n_i$ labeled examples (they are not assumed to be equal), and we assume that there is access to observed data from all tasks (or domains). In existing approaches for task and domain generalization, each task (or domain) is assumed to be a realization from a common unknown distribution, $\mathcal{T}_i \in P(\mathcal{T})$ (or $\mathcal{D}_i \in P(\mathcal{D})$), and they are expected to be related to each other for guaranteed success. However, this assumption is highly restrictive and often times the relationships between tasks or domains are not known \textit{a priori}. Hence, we develop a structured meta learning technique that infers the relationships between tasks/domains and simultaneously leverages this information to produce task/domain-specific models with improved performance. 
\vspace{5pt}

\noindent \textbf{Domain Design.} Domain shifts correspond to variations in the data statistics that can render a trained model ineffective~\cite{msda}, particularly when we do not do have access to data that is representative of the testing scenario. In this case, we assume that the $K$ datasets correspond to solving the same task, i.e. the same input and output label spaces, but are characterized by differences in the marginal feature distribution $P(\mathrm{X}_i)$ with identical conditional distributions $P(\mathrm{Y}_i|\mathrm{X}_i)$. Example domains that we consider in our experiments include a variety of image transformations such as scaling, color transformations, rotation etc.

\noindent \textbf{Task Design.} In this case, we assume each dataset corresponds to a binary classification problem of detecting the presence of a specific object, while the negative class is heterogeneous (contains images from multiple classes). However, we assume that there are no inherent domain shifts and the marginal feature distributions $P(\mathrm{X}_i)$ are identical.

\section{Background}
\label{sec:background}
\noindent \textbf{Handling Task/Domain Shifts:} Broadly, approaches for dealing with domain shifts can be categorized into domain adaptation~\cite{Hoffman2017CyCADACA, shu2018a,tzeng2017adversarial} and domain generalization~\cite{Li2017LearningTG} techniques. While the former adapts a pre-trained model using unlabeled or sparsely labeled target domain data, the latter is aimed at designing a model that can work well even in unseen target domains. When data from multiple domains are available at train time, one can utilize multi-domain learning methods~\cite{msda}, which attempt to extract domain-agnostic feature representations with the hope that these common factors can persist even with unseen test domains. More recently, model agnostic approaches that rely on meta-optimization (\textit{learning to learn})~\cite{Andrychowicz2016LearningTL} to improve the generalization of a base learner have gained a surge in interest. On the other hand, combating task shifts requires controlled knowledge transfer from a pre-trained model that was trained on a task related to the target task. When compared to conventional multi-task learning methods~\cite{ruder2017overview}, model agnostic meta-learning~\cite{Finn2017ModelAgnosticMF} have been found to be effective for few-shot learning scenarios.
\vspace{10pt}
 
\noindent \textbf{Model Agnostic Task/Domain Generalization:} Following~\cite{Li2017LearningTG,Finn2017ModelAgnosticMF}, we will now derive a generic model agnostic approach that applies to both task and domain generalization, which forms the core of \texttt{Invenio}. Though we develop the formulation for tasks $\{\mathcal{T}_i\}$, it is directly applicable to the case of domains as well. In order to enable the generalization of a unified model $\mathcal{F}$ with parameters $\Theta$ to all observed tasks, we first split the set of tasks into $K^{\dagger}$ meta-train and $K^{\ddagger} = K - K^{\dagger}$ meta-test tasks, $\mathcal{S}^{\dagger}$ and $\mathcal{S}^{\ddagger}$ respectively. 
 
Given the prediction $\hat{\mathrm{y}}$ for a sample $\mathrm{x}$ from task $\mathcal{T} \in \mathcal{S}^{\dagger}$, we can use the task-specific loss function $\ell({\mathrm{y}},\hat{\mathrm{y}})$ to measure its fidelity.
A typical meta-optimization strategy consists of two steps referred as meta-train and meta-test steps. In the meta-train step, the model parameters $\Theta$ are updated using the aggregated losses from the $K^{\dagger}$ meta-train tasks:
 
\begin{equation}
\mathcal{L}(\Theta) = \frac{1}{K^{\dagger}} \sum_{i=1}^{K^{\dagger}} \frac{1}{n_i} \sum_{(\mathrm{x},\mathrm{y}) \in (\mathrm{X}_i, \mathrm{Y}_i)} \ell\bigg({\mathrm{y}},\mathcal{F}(\mathrm{x}; \Theta)\bigg).
\label{eqn:tc_train}
\end{equation}This loss function is parameterized using $\Theta$ and hence the gradients are calculated with respect to this loss function, $\nabla_{\Theta} \mathcal{L}(\Theta)$. 
\begin{equation}
\hat{\Theta} = \Theta - \alpha  \nabla_{\Theta} \mathcal{L}(\Theta),
\end{equation}where $\alpha$ denotes the step size. In the meta-test step, the estimated parameters are evaluated on the $K^{\ddagger}$ meta-test tasks to virtually measure the generalization performance. Consequently, the aggregated loss function obtained using the updated parameters on the test tasks can be written as
\begin{equation}
\mathcal{G}(\hat{\Theta}) = \frac{1}{K^{\ddagger}} \sum_{j=K^{\dagger}+1}^{K} \frac{1}{n_j} \sum_{(\mathrm{x},\mathrm{y}) \in (\mathrm{X}_j, \mathrm{Y}_j)} \ell\bigg({\mathrm{y}},\mathcal{F}(\mathrm{x}; \hat{\Theta})\bigg).
\label{eqn:tc_test}
\end{equation}Our goal is to update the parameters $\Theta$ such that it can be effective for both meta-train and meta-test tasks. Hence, the overall objective is:
\begin{equation}
\arg \min_{\Theta} \mathcal{L}(\Theta) + \beta \mathcal{G}(\Theta - \alpha \nabla_{\Theta} \mathcal{L}(\Theta)).
\end{equation}To intuitively understand this objective, we follow the analysis in~\cite{Li2017LearningTG} and perform first-order Taylor expansion on the second term to obtain
$$\mathcal{G}(\Theta - \alpha \nabla_{\Theta} \mathcal{L}(\Theta)) = \mathcal{G}(\Theta) - \alpha \nabla_{\Theta} \mathcal{L}(\Theta) . \nabla_{\Theta} \mathcal{G}(\Theta),$$where the expansion is carried out around $\Theta$. Intuitively, the meta-optimization process amounts to minimizing the losses on training tasks while maximizing the dot product between the gradients from train and test tasks.
 
\section{Proposed Approach}
\label{sec:approach}
In this section, we present the proposed approach for inferring the semantic space of tasks or domains using only few-shot datasets. Without loss of generality, we set up the formulation for the task case, though it is applicable to domains as well. In contrast to existing meta-optimization approaches \cite{Finn2017ModelAgnosticMF} , we express the learner for each task $\mathcal{T}_i$ as a task-specific transformation $\mathcal{F}_i$ described using parameters $\Theta_i$, which maps input images from that task to its output label space $\mathrm{Y}_i$. Consequently, for a sample from the task $\mathcal{T}_i$, i.e., $\mathrm{x} \in \mathrm{X}_i$, the prediction can be obtained as: $\hat{\mathrm{y}} = \mathcal{F}_i(\mathrm{x}; \Theta_i).$ In our setup, we implement all these learners as deep networks (e.g convolutional networks). As described earlier, our formulation follows classical multi-task learning, where all datasets are assumed to be drawn from the same data distribution (no covariate shifts) and they differ only through task shifts~\cite{yang2017unifying}. Further, we assume that each task solves a binary classification problem, e.g. detecting the presence of a certain object in images. 

We now describe our approach that reveals the semantic relationships between tasks, and at the same time exploits this semantic similarity to produce improved generalization for all observed tasks. Though there are fundamental differences in how tasks and domains are defined in our setup (See Section \ref{sec:problem}, the meta-learning style optimization of \texttt{Invenio} is applicable to both tasks and domains. 

\RestyleAlgo{boxruled}
\begin{algorithm}[t]
	
	\KwIn{Set of tasks $\mathcal{T}$}
	\textbf{Initialization}: Parameters $\Theta_i$ for each task-specific model $\mathcal{F}_i$. Set hyper-parameters $\alpha, \beta, \gamma, \delta$ \;
	\textbf{Split}: $\mathcal{T}^{\dagger} \text{ and } \mathcal{T}^{\ddagger} \leftarrow \mathcal{T}$ \;
	\For{iter \textbf{in} $n_{iter}$}{
		/*\textit{Meta-Train Phase} \\
		\For {$\mathcal{T}_i$ \textbf{in} $\mathcal{T}^{\dagger}$}{
			Compute gradients $\nabla_{\Theta_i} \mathcal{L}_i(\Theta_i)$ \;
			Update $\hat{\Theta}_i = \Theta_i - \alpha  \nabla_{\Theta_i} \mathcal{L}_i(\Theta_i)$ \;
		}
		/*\textit{Meta-Test Phase} \\
		\For{$\mathcal{T}_i$ \textbf{in} $\mathcal{T}^{\dagger}$}{
			
			\For{$\mathcal{T}_j$ \textbf{in} $\mathcal{T}^{\ddagger}$}{
				$\eta_{ij} = \sum \nabla_{\Theta_i} \mathcal{L}_i(\Theta_i).\nabla_{\Theta_i} \mathcal{L}_j(\Theta_i)$ \;
			}
			Compute normalized scores $\bar{\eta}$ \;
			Compute meta-test loss $\mathcal{G}(\hat{\Theta}_i)$ using (\ref{eqn:ti_test_task}) \;
			/*\textit{Meta-optimization} \\
			Update $\Theta_i$:
			$$\Theta_i = \Theta_i - \delta \frac{\partial( \mathcal{L}_i(\Theta_i) + \beta \mathcal{G}(\hat{\Theta}_i)) }{\partial\Theta_i} \text{ };$$ 
		}
		$\mathcal{T}^{\dagger} \text{,} \mathcal{T}^{\ddagger}$ = $\mathcal{T}^{\ddagger} \text{,} \mathcal{T}^{\dagger}$ /* \textit{Swap meta-train/test sets}
	}
	\caption{Proposed structured meta-learning algorithm for the task case.}\label{algo}
	
\end{algorithm}
\vspace{10pt}

\noindent \textbf{Structured Meta-Learning with Task-Specific $\Theta_i$.} We follow the notations introduced in Section~\ref{sec:background}. The parameters for the task-specific transformations $\Theta_i, i=1,...,K$ are learned such that each $\mathcal{F}_i$ should generalize to semantically related tasks. We propose a novel structured meta-learning formulation to achieve this objective. During the meta-train step, for each of the train tasks $\mathcal{T}_i$, we compute the loss based on binary cross-entropy (cross-entropy for multi-class classification for domains) as follows:
\begin{equation}
\mathcal{L}_i(\Theta_i) = \frac{1}{n_i} \sum_{(\mathrm{x},\mathrm{y}) \in (\mathrm{X}_i, \mathrm{Y}_i)} \ell\bigg({\mathrm{y}},\mathcal{F}_i(\mathrm{x}; \Theta_i)\bigg).
\label{eqn:ti_train_task}
\end{equation}Here, the subscript for the loss $\mathcal{L}_i$ indicates that the data samples $\mathrm{X}_i$ from task $\mathcal{T}_i$ are used for evaluating the loss function with model parameters $\Theta_i$. The parameters $\Theta_i$ are then updated using the gradient update step as:
\begin{equation}
\hat{\Theta}_i = \Theta_i - \alpha  \nabla_{\Theta_i} \mathcal{L}_i(\Theta_i).
\end{equation}In order to ensure that the learned transformation $\mathcal{F}_i$ generalizes to related tasks, we need to first quantify the similarity between a meta-train task $\mathcal{T}_i$ and a meta-test task $\mathcal{T}_j$. More specifically, we measure the similarity between two tasks as the dot product between gradients with respect to the weights $\Theta_i$ relative to the losses evaluated on both the training domain data $\mathrm{X}_i$ and test domain data $\mathrm{X}_j$ respectively. Mathematically, this is expressed as:
\begin{equation}
\eta_{ij} = \sum \nabla_{\Theta_i} \mathcal{L}_i(\Theta_i).\nabla_{\Theta_i} \mathcal{L}_j(\Theta_i).
\label{eqn:ti_test_eta_task}
\end{equation}The summation in the above expression is over all parameters in the set $\Theta_i$, and the gradient estimate $\nabla_{\Theta_i} \mathcal{L}_i(\Theta_i)$ is obtained by summing over all mini-batches. Intuitively, input images from both tasks $\mathcal{T}_i$ and $\mathcal{T}_j$ are processed using the same transformation $\Theta_i$, and we expect the tasks to be related if the gradients for updating the parameters are in a similar direction. This intuition corroborates with existing formulations in~\cite{Task2Vec, achille2019information}, where it was shown that the gradients of weights of a neural network relative to a task-specific loss are a meaningful representation of the task itself. Hence the aggregated meta-test loss for updating the parameters $\Theta_i$ can be expressed as
\begin{equation}
\mathcal{G}(\hat{\Theta}_i) =  \sum_{j={K}^{\dagger}+1}^{K} \bar{\eta}_{ij} \frac{1}{n_j} \sum_{(\mathrm{x},\mathrm{y}) \in (\mathrm{X}_j, \mathrm{Y}_j)} \ell\bigg({\mathrm{y}},\mathcal{F}_j(\mathrm{x}; \hat{\Theta}_i)\bigg).
\label{eqn:ti_test_task}
\end{equation}Note, the $\eta_{ij}^{'s}$ for the set of meta-test tasks $\mathcal{T}_j^{'s}$ are normalized to sum to $1$ for obtaining $\bar{\eta}_{ij}^{'s}$. As it can be seen in the above expression, the similarity scores between tasks are used to determine which test tasks that the parameters $\Theta_i$ must generalize to and this naturally induces a semantic structure between the set of input tasks.
The overall objective for updating each $\Theta_i$ using the structured meta-optimization can thus be written as,
\begin{align}
\nonumber \arg \min_{\Theta_i} \mathcal{L}_i(\Theta_i) &+ \beta \mathcal{G}(\Theta_i - \alpha \nabla_{\Theta_i} \mathcal{L}_i(\Theta_i))
\label{eqn:obj_sm}
\end{align}

As mentioned above, the proposed strategy infers a semantic space of tasks or domains. This evolution of the semantic structure that occurs inherently as a part of our optimization strategy serves as a powerful tool in terms of its resourcefulness for generalization to new test scenarios, development of new tasks, investigating semantic similarities and unmasking new affiliations. 
A detailed algorithm of our approach is outlined in Algorithm \ref{algo}.

\section{Experiment: Task Shifts}
\label{sec:results_tasks}
In order to demonstrate \texttt{Invenio}, we construct a custom task  database, study the inferred semantic structure, and perform quantitative evaluation in terms of classification performance across all tasks. Our database is a collection of $400$ tasks sampled from four different benchmark image classification datasets, with the underlying assumption that there are no inherent covariate shifts. As described earlier, our focus is on the few-shot learning scenarios and we do not assume prior knowledge of the underlying relationships between the constituent tasks. 

\begin{figure}[t]
	\centering
	\includegraphics[width=0.95\linewidth]{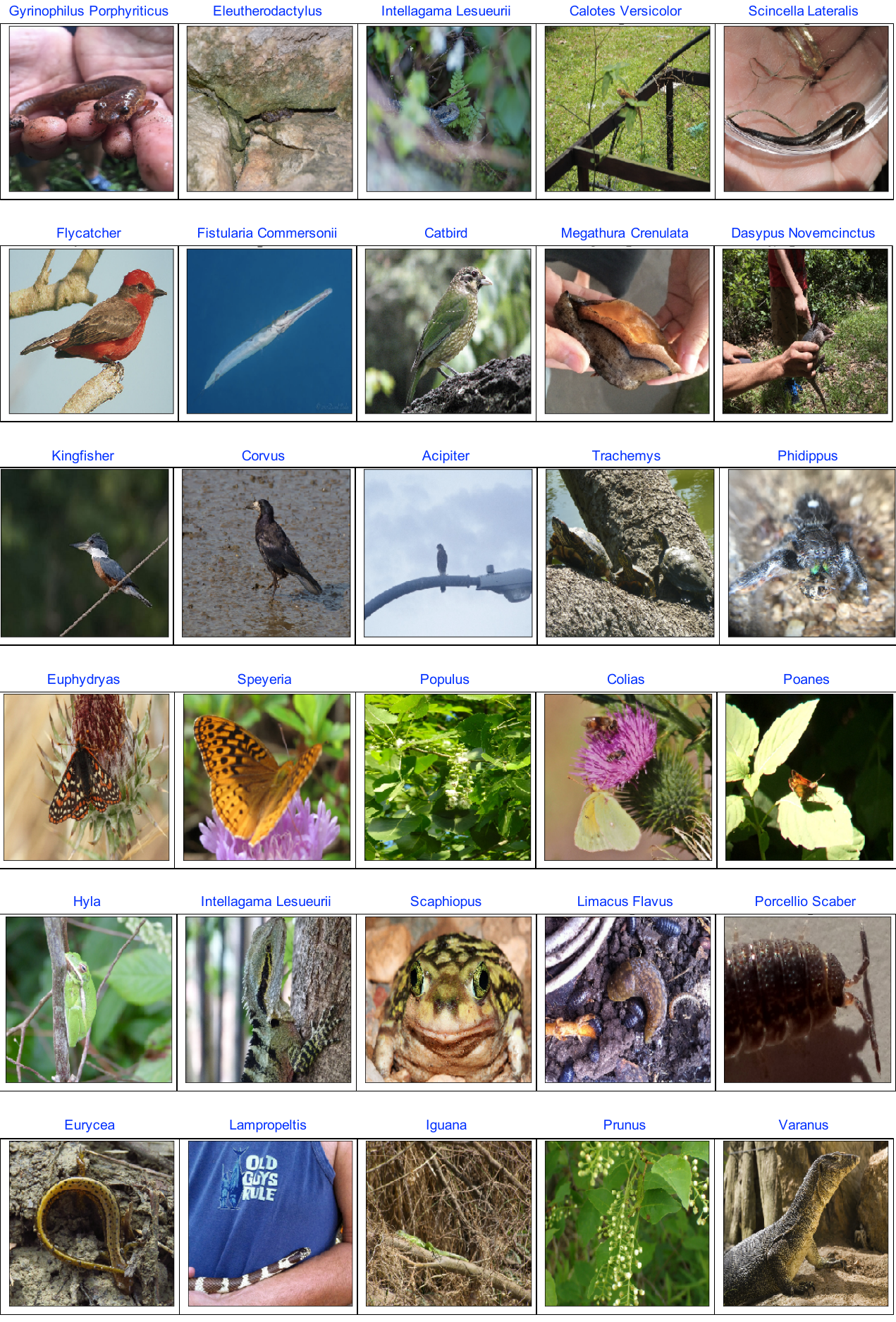}
	\caption{Semantic space of tasks - \texttt{Invenio} reveals interesting relationships between tasks. Here we show the most similar tasks for each query task (leftmost in each row).}
	\label{fig:task_nn_ex}
\end{figure}
\vspace{-5pt}
\begin{figure*}[t]
	\centering
	\subfloat[2D Embedding]{\includegraphics[width=0.45\textwidth,valign=c]{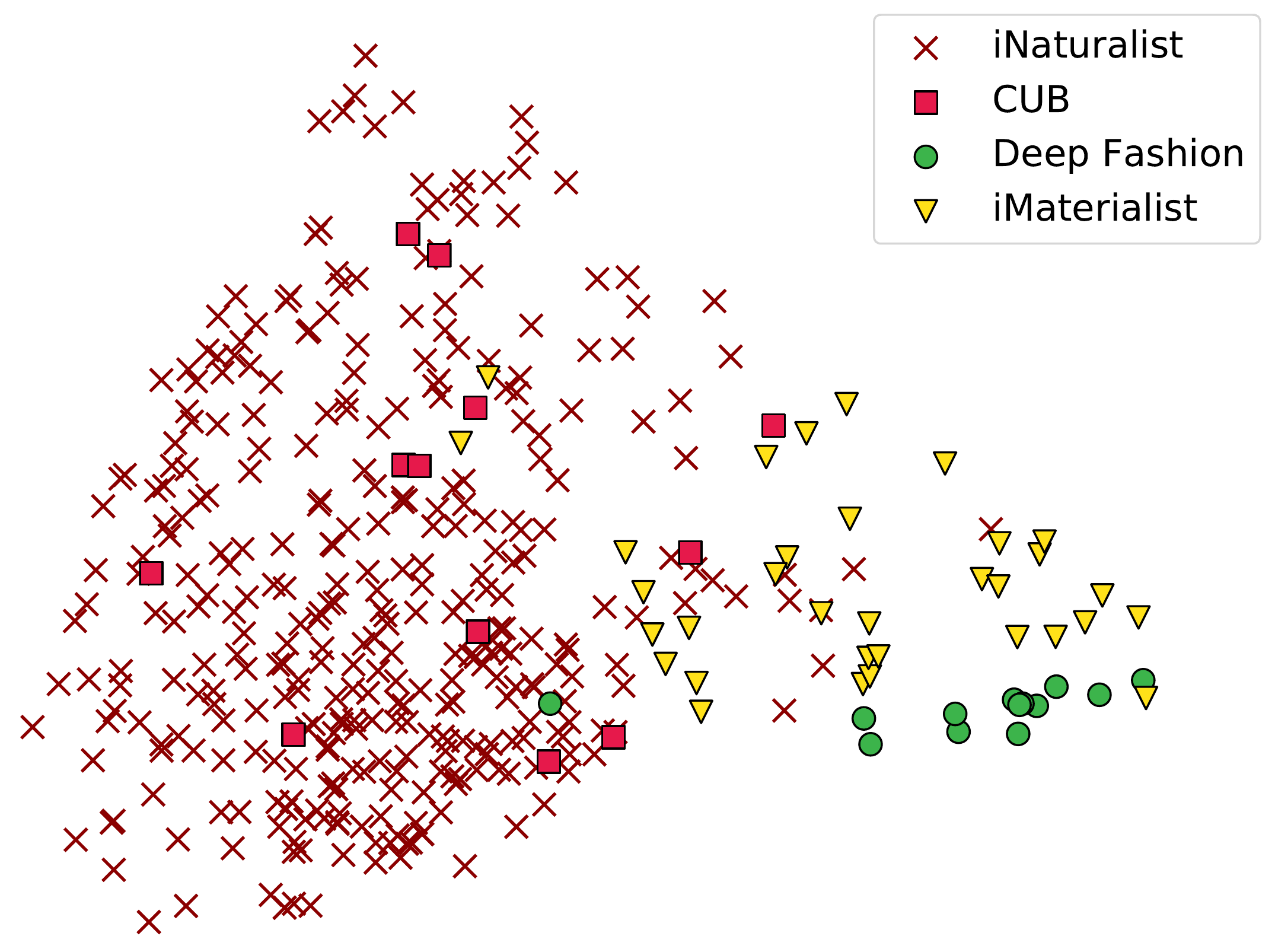}}
	\hspace{0.1in}
	\subfloat[Performance Evaluation]{\includegraphics[width=0.47\textwidth,valign=c]{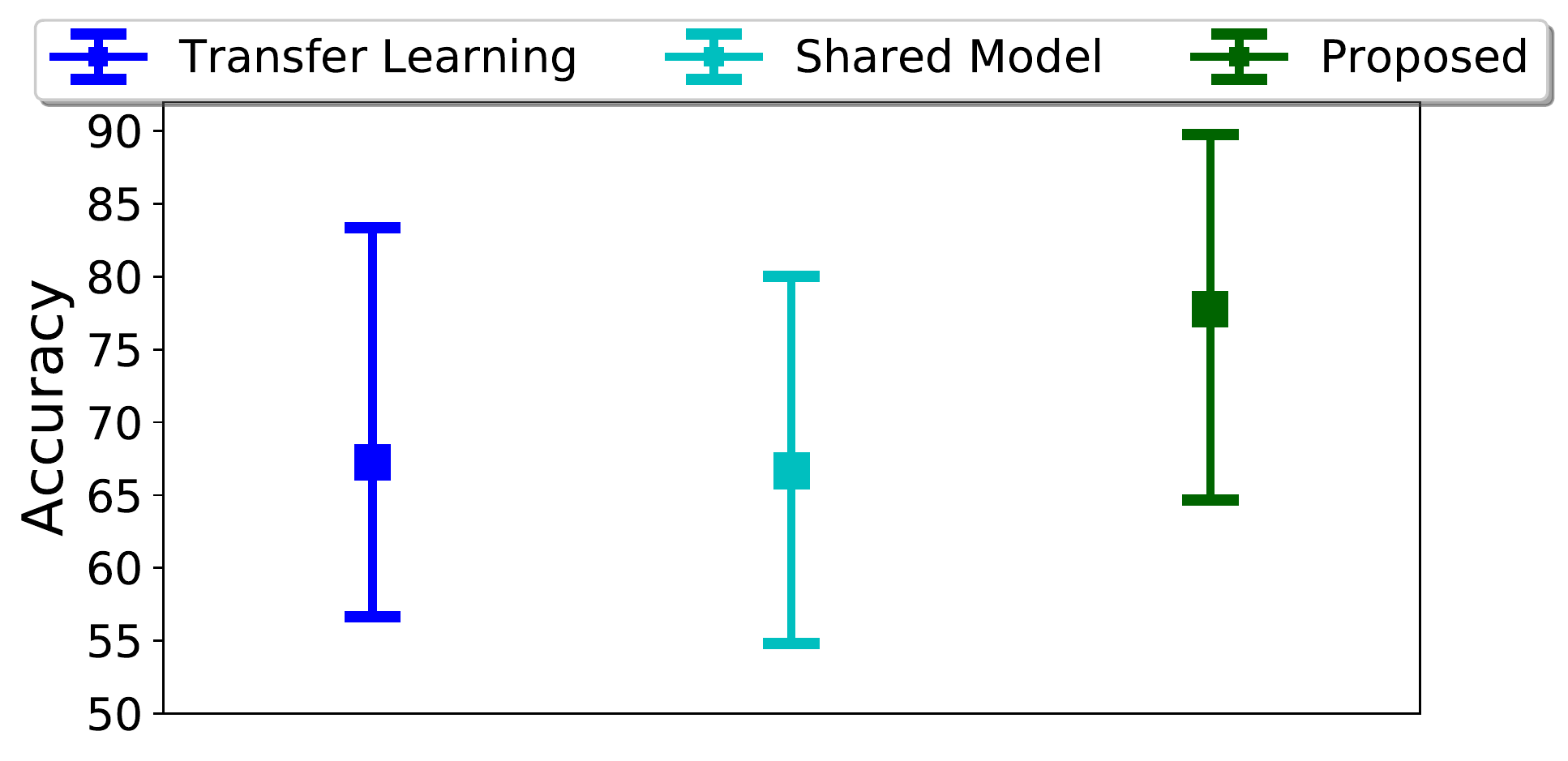}}
	\hfill
	\subfloat[Comparison to Transfer Learning]{\includegraphics[width=0.95\textwidth]{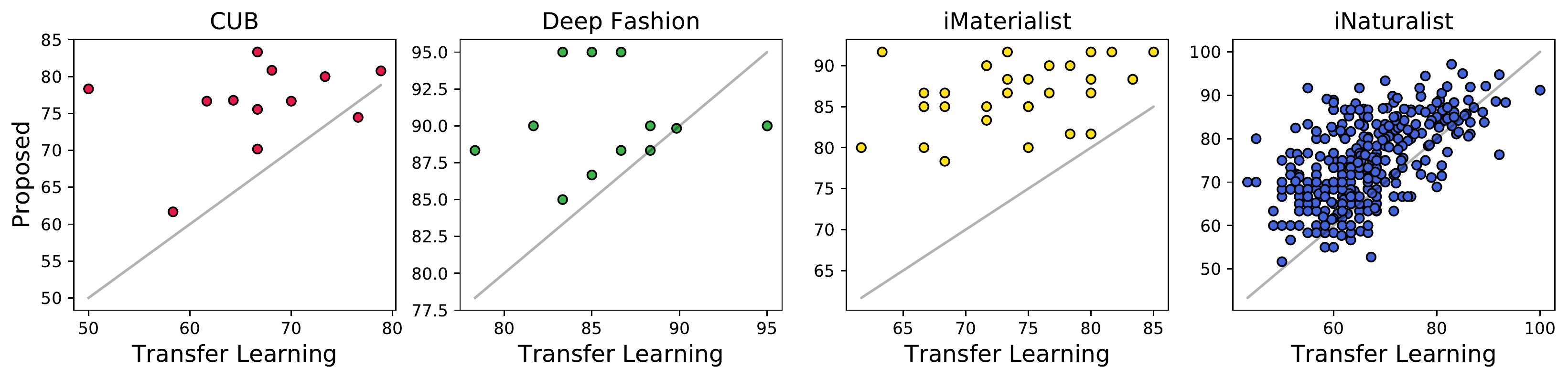}}
	\caption{Evaluating \texttt{Invenio} on the custom task database - (a) $2$D visualization of the inferred semantic structure, (b) Median, along with $25^{\text{th}}$ and $75^{\text{th}}$ quantiles, of the test accuracies for the entire task database, (c) Fine-grained evaluation of the task-specific accuracies in comparison to transfer learning from a common model.}
	\label{fig:task_comparison}
	\vspace{-5pt}
\end{figure*}
\vspace{10pt}

\noindent \textbf{Task Database Design.} Our task database consists of $400$ binary classification tasks sampled from four different datasets namely CUB~\cite{CUB} , DeepFashion~\cite{deepfashion}, iMaterialist~\cite{imaterialist} and iNaturalist~\cite{inaturalist}. While the positive class in each task corresponds to a specific image class from one of the datasets, the negative class contains images (randomly chosen) from all datasets.

\noindent\textit{CUB200}~\cite{CUB}: The Caltech-UCSD Birds dataset contains a total of $6000$ images from $200$ different bird species. We randomly selected $12$ categories and included them as $12$ different binary classification tasks.

\noindent\textit{Deep Fashion}~\cite{deepfashion}: This is a large-scale clothes database with diverse images from $50$ categories. We used images from $13$ randomly selected categories to construct our tasks.

\noindent\textit{iMaterialist}~\cite{imaterialist}: We considered $33$ categories from this large-scale fashion dataset to form our tasks. Note that, this curation was performed such that some of the chosen categories overlap with those selected from Deep Fashion.

\noindent\textit{iNaturalist}~\cite{inaturalist}: This is a large-scale species detection dataset. We sampled $342$ categories from broad taxonomical classes such as Mammalia, Reptalia, Aves etc. Note, in the selected set of $342$ tasks, $69$ tasks correspond to birds from the Aves category, and hence we expect to observe semantic relevance to tasks from the CUB database.

By design, there is partial overlap in tasks between iNaturalist and CUB datasets, and similarly between iMaterialist and DeepFashion, while simultaneously there is a clear disconnect between fashion and species datasets. Such a design enables us to evaluate our approach and reason about the discovered semantic structure between tasks. Each binary classification problem contains a maximum of $100$ positive samples (from a specific image class), while another $100$ randomly chosen samples from the remaining set of $399$ tasks constitute the negative class.
\vspace{10pt}

\noindent \textbf{Architecture.} We use the same model architecture for each task $\mathcal{T}_i$, but with individual parameter sets $\Theta_i$ -- {Conv(3,20,3,3), ReLU, MaxPool, Conv(20,50,3,3), ReLU, MaxPool, Linear(2450,500), ReLU, Linear(500,1), ReLU}. Note, we resize all images to $128 \times 128$ pixels. The learning rates for the meta-train and meta-test phases were set to $1$e$-4$ and $1$e$-3$ respectively.

\begin{table}[t]
	\centering
	\caption{Effect of meta-test batch size on the classification performance across the entire set of tasks.}
	\vspace{5pt}
	\renewcommand*{\arraystretch}{1.2}
	\begin{tabular}{|c|c|c|c|}
		\hline
		\multirow{2}{*}{\begin{tabular}[c]{@{}c@{}}\textbf{Test}\\ \textbf{Accuracy}\end{tabular}} & \multicolumn{3}{c|}{\textbf{Meta-Test Batch Size}}                                                   \\\cline{2-4}
		& \textbf{5 }                        & \textbf{12}                        & \textbf{20}                                 \\
		\hline \hline
		Median                                                                   & 74.57                     & 75.54                     & \textbf{75.98}                     \\
		$25^{\text{th}}$ Quantile                                                & 66.67                     & 68.33                    & \textbf{69.33}                              \\
		$75^{\text{th}}$ Quantile                                                & 81.67 & 83.33 & \textbf{84.21} \\
		\hline
	\end{tabular}
	\label{table:batch}
	\vspace{-15pt}
\end{table}
\vspace{10pt}

\noindent \textbf{Semantic Space of Tasks.} \texttt{Invenio} jointly infers the inherent semantic structure and optimizes the task-specific model parameters through a structured meta-learning approach. In order to analyze the inferred semantic space, upon completion the training process, we compute the pairwise task similarities between all $400$ tasks using \eqref{eqn:ti_test_eta_task}. Denoting the similarity matrix by $\mathrm{S} \in \mathbb{R}^{400 \times 400}$, we perform truncated SVD to obtain low-dimensional embeddings for analysis and visualization of the task relationships. Figure~\ref{fig:task_comparison}(a) illustrates a $2$D visualization of the task space, which clearly reveals a separation between the fashion and species datasets. In Figure~\ref{fig:task_nn_ex}, we show the most similar tasks for each query task (left most in each row) in the embedding space ($8$D embeddings). We find that the semantics inferred by \texttt{Invenio} matches human knowledge in these examples, and hence can be expected to lead to improved generalization when solved jointly. For example, the \textit{Formal Dresses} task is found to be semantically similar with other types of dresses, \textit{Tees}, \textit{Cardigans} etc., while \textit{Eurycea} (snake) is in the neighborhood of other reptiles, .e.g. \textit{Iguana} and \textit{Varanus}. However, we also find somewhat unexpected relationships -- \textit{Flycatcher} and \textit{Fistularia Commersonii} (fish), or \textit{Euphydryas} (butterfly) and \textit{Populus} (plant) due to the occurrence of similar visual patterns though they are semantically unrelated.

\begin{figure}[t]
	\centering
	\includegraphics[width=0.95\linewidth]{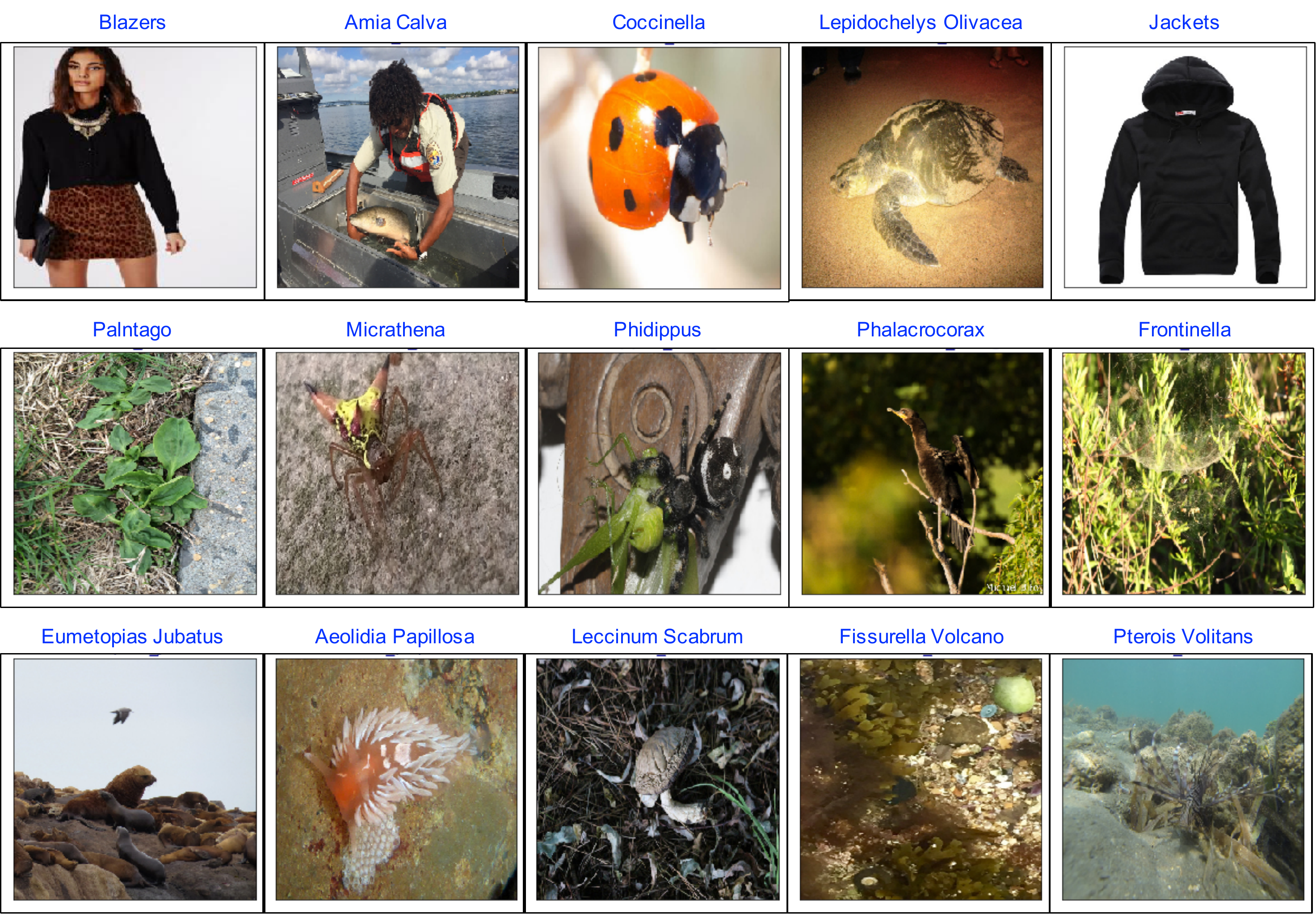}
	\caption{Examples cases where the relationships identified are not easily interpretable, but still leads to improved classification performance.}
	\label{fig:task_nn_fail_ex}
	\vspace{-15pt}
\end{figure}

\noindent \textbf{Performance Evaluation:} Our hypothesis is that by leveraging the inferred semantic structure into the learning process, we can improve the quality of the task-specific predictive models. To this end, we evaluate the classification performance of \texttt{Invenio} on a held-out test set for each of the $400$ tasks in the database. For comparison, we consider the two popular baselines: (i) \textit{Transfer Learning:} This is the most commonly adopted strategy for task adaptation. We train a model, with the same architecture as ours on the complete CIFAR10 dataset~\cite{cifar10} and subsequently fine-tune the model using labeled data from each of the tasks independently; and (ii) \textit{Shared Model:} This approach assumes a shared model across all the tasks and employs the model agnostic meta leanring (MAML)~\cite{Finn2017ModelAgnosticMF} technique to optimize for the model parameters such that it generalizes to the entire set of tasks. We use the accuracy metric to measure the performance and we report results on the held-out test set for all cases. Note that, the proposed approach can be viewed as a generalization of the \textit{Shared Model} baseline, wherein the task relationships are exploited while updating the task-specific model parameters.

Figure~\ref{fig:task_comparison}(b) compares the classification performance of the different approaches on the entire task database. In particular, we show the median, along with $25^{\text{th}}$ and $75^{\text{th}}$ quantiles, of test accuracies across all tasks. As evidenced from the plot, by exploiting the semantic structure of the space of tasks, the proposed approach significantly outperforms the baseline methods. A fine-grained evaluation of \invenio in comparison to \textit{Transfer Learning} for each of the tasks can be found in Figure~\ref{fig:task_comparison}(c). A critical parameter of the proposed approach in Algorithm~\ref{algo} is the meta-test batch size. While utilizing the entire meta-test set $\mathcal{T}^{\ddagger}$ would allow us to identify the relationships effectively, it is not computationally feasible. Hence, in practice, we use a smaller batch size. From Table~\ref{table:batch}, we find that, increasing the batch size until from $5$ to $12$ leads to appreciable improvements, while we observe no significant improvements beyond $20$.

While the examples showed in Figure~\ref{fig:task_nn_ex} reasonably agree with our understanding of task similarities, we also find cases (see Figure~\ref{fig:task_nn_fail_ex}) where the relationships are not easily justified. Nevertheless, the inferred structure still provided significant performance gains. For example, for the \textit{Blazers} task, the neighborhood is highly diverse, however \texttt{Invenio} achieves $88\%$ test accuracy compared to $78\%$ with transfer learning. Similarly, images from the \textit{Eumetopias Jubatu} task (sea lion) contains complex visual patterns leading to not-so-interpretable relationships. Surprisingly, this achieves a large performance gain of $~24\%$ over transfer learning. These results clearly evidence the importance of considering similarities between tasks to build highly effective predictive models. 

\section{Experiment: Domain Shifts}
\label{sec:results_domains}
The model agnostic nature of \invenio allows its use in revealing hidden relationships between datasets that involve complex covariate shifts. To this end, we develop a custom domain database and demonstrate the effectiveness of \invenio in improving the fidelity of the resulting domain-specific classifiers. 
\begin{figure*}[t]
	\centering
	\includegraphics[width=0.90\textwidth]{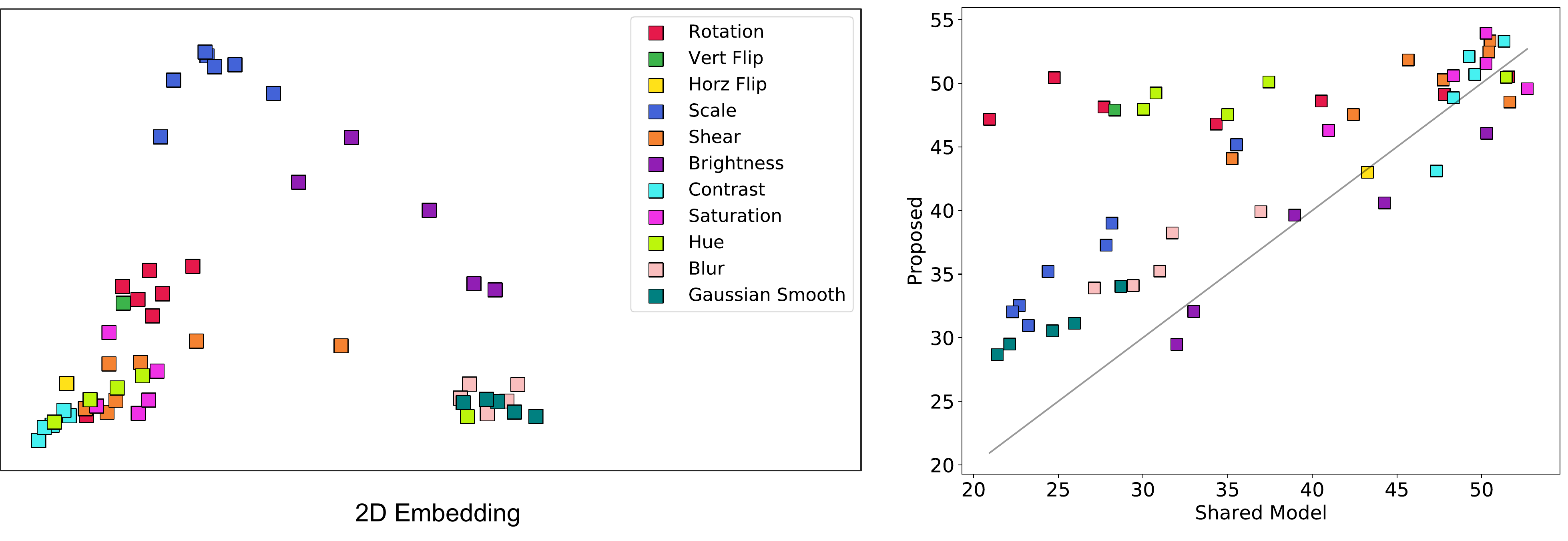}
	\caption{Evaluating \texttt{Invenio} on the custom domain database - (a) $2$D visualization of the inferred semantic structure, (b) Fine-grained evaluation of the domain-specific accuracies in comparison to learning with a shared model.}
	\label{fig:domain}
	\vspace{-10pt}
\end{figure*}

\begin{figure}[t]
	\centering
	\includegraphics[width=0.65\linewidth]{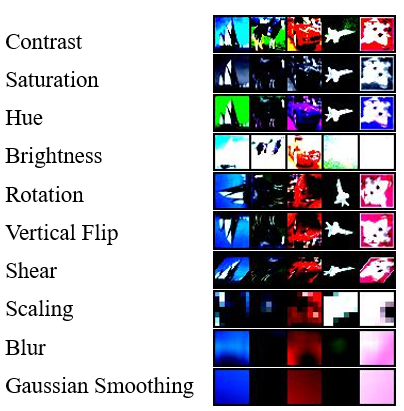}
	\caption{Example images for each of the domain shifts considered in our setup.}
	\label{fig:domain_ex}
	\vspace{-10pt}
\end{figure}
\vspace{10pt}

\noindent \textbf{Domain Database Design:} Our domain database is composed $53$ different variants of the CIFAR-10 dataset~\cite{cifar10}, obtained using a broad class of image transformations, while solving the same task of multi-class classification ($10$ classes). Here is the complete list of domain shifts considered: (i) \textit{Rotation}: $7$ variants were generated by rotating the images, where the degree of rotation was varied between $0$ to $90$; (ii) \textit{Flip}: We generated $2$ datasets by applying horizontal and vertical flips to the images. These transformations can be viewed as special cases of \textit{Rotation}; (iii) \textit{Affine}: We constructed $14$ domains by applying different affine transformations to images and this was carried out by varying the settings for scale and shear; (iv) \textit{Color}: $20$ different datasets were created by manipulating parameters pertinent to color transformations, namely brightness, saturation, contrast and hue; and (v) \textit{Filter}: We used blurring and Gaussian smoothing techniques to create $10$ variants of the base domain. While Gaussian smoothing produces blurring by applying Gaussian function based transformation on image pixels, the Box Blur filter replaces each pixel by the average of its neighboring pixels.

Intuitively, we expect geometric transformations such as \textit{Affine}, \textit{Rotation} and \textit{Flip} to be related among themselves and can benefit by shared feature representations. On the other hand, transformation such as hue, saturation, contrast and brightness are expected be strongly related. Each domain is comprised of 300 randomly chosen samples from each class and the performance evaluation is carried out using a held-out test set for all $53$ domains.
\vspace{10pt}

\noindent\textbf{Architecture:} For all the domain specific base learners, we use the same architecture --
{Conv(3,20,5,1), ReLU, MaxPool, Conv(20,50,5,1), ReLU, MaxPool, Linear(2450,500), ReLU, Linear(500,10), ReLU} which follows the same syntax as tasks. Similar to the previous experiment, the learning rates for the meta-train and meta-test phases were set to $1$e$-4$ and $1$e$-3$ respectively.
\vspace{10pt}

\noindent \textbf{Semantic Space of Domains.} Figure~\ref{fig:domain}(a) provides a $2$D visualization of the semantic space obtained by applying truncated SVD on the similarity matrix $\mathrm{S} \in \mathbb{R}^{53 \times 53}$ between the set of domains. As it can be observed, the structure largely aligns with our hypothesis, i.e., the geometric transforms such as, rotation, flip and shear are closely related to each other. An interesting outcome is that the scale transformation does not belong in the same part of the semantic space as the other geometric transformations. We attribute this to the information loss that occurs due to cropping of the zoomed image to remain within the original boundaries.

Similar observations can be made about domains constructed based on color transformations to the original images. It is evident from Figure.~\ref{fig:domain}(a) that the datasets generated by manipulating hue, saturation and contrast respectively, are closely related to each other. However, brightness changes  manifest as being completely unrelated to other standard color transformations. As illustrated in Figure~\ref{fig:domain_ex}, this may be partly due to the high degree of brightness change that we applied, which caused the shadows/darker regions to mask the crucial features like edges. On the other hand the \textit{Contrast} transformation makes separation between dark and bright regions more prominent. Finally, the two filtering transformations that we considered are found to carry shared knowledge about the images, since both of them produce low-pass variants of the original images. 

\noindent \textbf{Performance Evaluation.} Similar to the task shifts case, we evaluate the effect of incorporating the learned semantic relationships into the domain-specific model optimization, in terms of performance on held-out test data. In Figure~\ref{fig:domain}(b), we compare the test accuracies for all $53$ domains against the baseline approach where all $53$ domains use a shared model. We find that \texttt{Invenio} produces significant performance improvements particularly in the cases of \textit{Rotation}, \textit{Hue} and \textit{Shear} transformations, which actually corresponds to the densest part of the semantic space. This indicates that \textit{Invenio} is able to perform meaningful data augmentation, thus leading to models with higher fidelity.

\section{Conclusions}
\label{sec:discussion}
In this paper, we introduced~\texttt{Invenio}, a structured meta learning approach that infers the inherent semantic structure, and provides insights into the complexity of transferring knowledge between different tasks (or domains). Unlike existing approaches such as \texttt{Task2Vec}~\cite{Task2Vec} and Taskonomy~\cite{Zamir2018TaskonomyDT}, which  compare tasks by measuring similarities between pre-trained models, \texttt{Invenio} adopts a self-supervised strategy by identifying the semantics inside the training loop. Furthermore, our approach is applicable to few-shot learning settings and can scale effectively to a large number of tasks. Finally, the inferred semantics largely agree with our intuition, and even when they do not, they still help in improving the classification performance over existing transfer learning strategies.

An important outcome of this work is that the insights from \texttt{Invenio} can be utilized to produce powerful task/domain priors, which can in turn be used to sample new tasks, akin to generative models for data. This work belongs to the class of recent approaches that are aimed at abstracting learning objectives in AI systems~\cite{milli2017interpretable, liu2019self}. Designing effective strategies for sampling from these task priors, and building algorithmic solutions for data augmentation remain part of future work. 

{\small
\bibliographystyle{ieee_fullname}
\bibliography{invenio}
}

\end{document}